\documentclass[conference]{IEEEtran}
\IEEEoverridecommandlockouts
\usepackage{cite}
\usepackage{amsmath,amssymb,amsfonts}
\usepackage{algorithmic}
\usepackage{graphicx}
\graphicspath{ {./images} }
\usepackage{wrapfig}
\usepackage{subcaption}
\usepackage{textcomp}
\usepackage{hyperref}
\usepackage{makecell}
\usepackage[usenames,dvipsnames]{color}
\hypersetup{
    colorlinks=true,
    linkcolor=blue,    % Color of internal links
    citecolor=blue,   % Color of citations
    filecolor=magenta, % Color of file links
    urlcolor=blue,     % Color of external URLs
    pdfborder={0 0 0}  % Removes the border around links
}

\usepackage{float}
\usepackage{xcolor}
\usepackage{url}

\usepackage{multirow}
\usepackage{array}
\usepackage{booktabs}
\usepackage{subcaption}
\newcommand{\PreserveBackslash}[1]{\let\temp=\\#1\let\\=\temp}
\newcolumntype{C}[1]{>{\PreserveBackslash\centering}p{#1}}
\newcolumntype{R}[1]{>{\PreserveBackslash\raggedleft}p{#1}}
\newcolumntype{L}[1]{>{\PreserveBackslash\raggedright}p{#1}}
\def\BibTeX{{\rm B\kern-.05em{\sc i\kern-.025em b}\kern-.08em
    T\kern-.1667em\lower.7ex\hbox{E}\kern-.125emX}}
\begin{document}

% A comprehensive benchmark for Bengali Handwitten Character and Digit Recognition using Few-shot Learning.
\title{Performance Analysis of Few-Shot Learning Approaches for Bangla Handwritten Character and Digit Recognition\\
}

% \author{(Anonymous STI Submission)}
%%%%%%% Author Info %%%%%%%%%%%%%
\author{ 
    \IEEEauthorblockN{Mehedi Ahamed, Radib Bin Kabir, Tawsif Tashwar Dipto, \\Mueeze Al Mushabbir, Sabbir Ahmed, and Md. Hasanul Kabir\\}

    \IEEEauthorblockA{Department of Computer Science and Engineering, Islamic University of Technology, Gazipur, Bangladesh\\}
    
    \IEEEauthorblockA{Email: \{mehediahamed, radib, tawsiftashwar, almushabbir, sabbirahmed, hasanul\}@iut-dhaka.edu}
}

%%%%%% For Header %%%%%%%%%%%%

\makeatletter
\let\old@ps@IEEEtitlepagestyle\ps@IEEEtitlepagestyle
\def\confheader#1{%
    % for the first page
    \def\ps@IEEEtitlepagestyle{%
        \old@ps@IEEEtitlepagestyle%
        \def\@oddhead{\strut\hfill#1\hfill\strut}%
        \def\@evenhead{\strut\hfill#1\hfill\strut}%
    }%
    \ps@headings%
}
\makeatother

\confheader{
    \parbox{\textwidth}{%
        \raggedleft
        2024 6th International Conference on Sustainable Technologies for Industry 5.0 (STI)\\
        14-15 December, Dhaka
    }
}

\IEEEpubid{
\begin{minipage}[t]{\textwidth}\ \\[10pt]
      \small{979-8-3315-3197-3/24/\$31.00 
      \copyright2024 IEEE 
      }
\end{minipage}
}

% \author{\IEEEauthorblockN{Tawsif Tashwar Dipto}
% \IEEEauthorblockA{\textit{Computer Science and Engineering} \\
% \textit{Islamic University of Technology}\\
% Dhaka, Bangladesh \\
% ID: 200041105 \\
% tawsiftashwar@iut-dhaka.edu}
% \and
% \IEEEauthorblockN{Radib Bin Kabir}
% \IEEEauthorblockA{\textit{Computer Science and Engineering} \\
% \textit{Islamic University of Technology}\\
% Dhaka, Bangladesh \\
% ID: 200041122 \\
% radib@iut-dhaka.edu}
% \and
% \IEEEauthorblockN{Mehedi Ahamed}
% \IEEEauthorblockA{\textit{Computer Science and Engineering} \\
% \textit{Islamic University of Technology}\\
% Dhaka, Bangladesh \\
% ID: 200041125 \\
% mehediahamed@iut-dhaka.edu}
% }

\maketitle

\begin{abstract}
Few-shot learning (FSL) offers a promising solution for classification tasks with limited labeled examples, offering a valuable solution for languages with limited annotated samples.
Traditional deep learning research has largely centered on optimizing performance using large-scale datasets, yet constructing extensive datasets for all languages is both labor-intensive and impractical. FSL offers a compelling alternative, achieving effective results with minimal data. In this connection, this study investigates the performance of FSL approaches in Bangla characters and numerals recognition with limited labeled data, demonstrating their applicability to scripts with intricate and complex structures where dataset scarcity is prevalent. Given the complexity of Bangla scripts, we posit that models capable of performing well on these characters will generalize effectively to languages of similar or lower structural complexity.
We introduce SynergiProtoNet, a hybrid network designed to enhance the recognition accuracy of handwritten characters and digits. Our model combines advanced clustering methods with a robust embedding framework to capture fine-grained details and contextual subtleties, leveraging multi-level (high- and low-level) feature extraction within a prototypical learning framework. We rigorously benchmark SynergiProtoNet against several state-of-the-art few-shot learning models, including BD-CSPN, Prototypical Network, Relation Network, Matching Network, and SimpleShot, across diverse evaluation settings. Our experiments--— Monolingual Intra-Dataset Evaluation, Monolingual Inter-Dataset Evaluation, Cross-Lingual Transfer, and Split Digit Testing demonstrate that SynergiProtoNet consistently achieves superior performance, establishing a new benchmark in few-shot learning for handwritten character and digit recognition. 
% reducing the requirement for extensive data collection and 
% Our model reduces the requirement for extensive data collection for training, which is expensive and resource-demanding. Its capacity to efficiently adapt to new data with minimal retraining upholds the fundamental principles of sustainable technology essential for modern industries.
%This advancement enhances automated OCR and error reduction in key sectors like banking, insurance, and government services.% 
The code is available on GitHub: \href{https://github.com/MehediAhamed/SynergiProtoNet}{https://github.com/MehediAhamed/SynergiProtoNet}.

\end{abstract}


\section{Introduction}

% Establishing Bangla Recognition as challenging and necessary research field
Automated handwritten character and digit recognition systems are integral to Optical Character Recognition (OCR) applications and document analysis \cite{0025Ocr, 0027documentAnalysis}. Bangla, ranked as the seventh most spoken language globally with over 270 million speakers, \cite{aziz2023banglaSER, yasmeen2021csvcNet}, where the script consists of 50 characters (11 vowels and 39 consonants), numerous compound characters, and 10 numerals, is both intricate and challenging for automated recognition. The variability in handwriting styles \cite{0033banglavariability} and the high similarity between certain characters further complicate the task of accurate recognition \cite{rahman2022twoDecades,ahmed2019aComplete, talha2017bangla}. These challenges are particularly relevant in sectors such as education, banking, and government services, where reliable recognition is essential for digitizing documents, verifying identity records, and processing financial transactions. 

Despite such solutions being relevant for all languages, their development is often hindered owing to the scarcity of large-scale datasets across different scripts. Although Deep Learning (DL) based solutions have achieved state-of-the-art performance in a wide variety of computer vision tasks owing to their powerful ability to learn complex and abstract features \cite{khatun2021signLanguage, ahmed2022less, alamgir2022performance, ashik2022recognizing}, most of the conventional CNN and DL-based models proposed for handwritten character and digit recognition \cite{0005dlbangla, 0003banglanet_Saha_Rahman_2024a, 0004bornonet, kamal2022huruf} require large-scale labeled datasets, which are often unavailable for low-resource languages. Few-shot learning (FSL) has emerged as a promising technique to overcome these hurdles by enabling models to learn effectively from limited data, thus improving the accuracy and efficiency of recognition systems across industries \cite{0030FSLsurvey,ahmed2022classification, rahman2024fusednetenhancingfewshottraffic}. By learning from a few examples, FSL mitigates the need for large datasets, accelerating the development of effective recognition systems for underrepresented languages.

Amongst the handful of recent works in handwritten character recognition using FSL models, Sahay and Coustaty \cite{0020urdufsl} leveraged a technique for Urdu handwritten character recognition by addressing its bi-directional nature, while Samuel et al. \cite{0018samuel2022offlinehandwrittenamhariccharacter} applied similar techniques to Amharic scripts, capturing its unique features. 
For Tamil script, Shaffi and Hajamohideen \cite{0019tamil} combined CNNs with RNNs to enhance recognition rates. FSL has also shown promise in low-resource languages like Persian \cite{0021persian} and in tasks like signature verification \cite{0022siamesesignature}, with novel models proving effective in few-shot scenarios \cite{0023novelsiamese}. These advancements highlight the potential of few-shot learning models in improving handwritten character recognition across various scripts, however, a thorough analysis of the recent state-of-the-art FSL-based models is yet to be investigated in the literature. 

This study presents a critical performance analysis of FSL approaches for handwritten Bangla characters and digit recognition. We introduce SynergiProtoNet, a hybrid model that combines a CNN encoder and a ResNet18 backbone to achieve a rich and diverse feature representation. The CNN encoder captures low-level features such as edges and textures, while the ResNet18 component extracts high-level, complex features, resulting in a model that excels in few-shot learning tasks. SynergiProtoNet sets a new benchmark in this task by addressing the complexities of the Bengali script and leveraging FSL for superior recognition accuracy and efficiency.

The proposed solution reduces the requirement for extensive data collection, which is expensive and resource-demanding. Adoption of our model in industry applications can improve the performance of automatic information retrieval from handwritten documents, streamlining workflows and reducing manual data entry errors in sectors such as banking, insurance, and government services, hence contributing to the big picture of achieving sustainable technology for the modern era.

% \section{Related Work}
% Recent advancements in neural network architectures, such as Convolutional Neural Networks (CNNs) and specialized models like BanglaNet \cite{0003banglanet_Saha_Rahman_2024a}, Bornonet \cite{0004bornonet}, and Deep Neural Networks (DNNs) \cite{0005dlbangla}, have significantly improved handwritten character recognition. CNNs excel in this domain by learning hierarchical features, which has been particularly effective for scripts like Bangla \cite{0016Ashikur_Rahman_2022}. Models like BanglaNet and Bornonet are specifically designed to handle the complexities of Bangla script, enhancing accuracy by addressing character similarity and optimizing network efficiency \cite{0003banglanet_Saha_Rahman_2024a, 0004bornonet}.

% In other scripts, FSL has emerged as a key technique. Sahay and Coustaty \cite{0020urdufsl} advanced Urdu handwritten character recognition by addressing its bi-directional nature, while Samuel et al. \cite{0018samuel2022offlinehandwrittenamhariccharacter} applied similar techniques to Amharic, capturing its unique features. For Tamil script, Shaffi and Hajamohideen \cite{0019tamil} combined CNNs with RNNs to enhance recognition rates. FSL has also shown promise in low-resource languages like Persian \cite{0021persian} and in tasks like signature verification \cite{0022siamesesignature}, with novel models proving effective in few-shot scenarios \cite{0023novelsiamese}. These advancements highlight the potential of few-shot learning models in improving handwritten character recognition across various scripts.

\section{Methodology}
In this study, we leveraged Few-Shot Learning (FSL) to tackle the challenge of recognizing Bangla handwritten characters and digits with limited data. FSL enables generalization from limited examples through the use of a `support set'--- a small labeled dataset that provides a few representative samples per class, and a `query set'--- which includes unlabeled examples for classification after the model has been exposed to the support set. A novel class refers to categories that the model encounters for the first time during testing.

\subsection{Datasets}
To ensure a thorough performance analysis of the FSL models under different evaluation methods, we have leveraged four datasets namely BanglaLekha-Isolated \cite{0006banglalekhaisolated}, CMATERdb 3.1.2 \cite{0007md__mostafizur_rahman_2023_cmaterdb}, Devanagari dataset \cite{0024Devanagari}, and NumtaDB \cite{0015numtadb}. %Table \ref{tab:dataset images} summarizes the sample counts of these datasets. 

BanglaLekha-Isolated is a comprehensive dataset containing approximately 110,000 images across 50 Bangla characters and 10 numerals, collected from a diverse pool of writers, capturing significant variations in handwriting styles. 
CMATERdb 3.1.2 provides 300 images per class and includes characters from Bangla, Devanagari, and Tamil scripts, offering a multi-script dataset for evaluating cross-lingual performance.
NumtaDB is focused on Bengali handwritten digits, with contributions from over 2,700 writers, encompassing various transformations and conditions to capture real-world variability. 
The Devanagari dataset includes 36 characters and 10 numerals, comprising 78,200 images in the training set and 13,800 images in the test set.

\subsection{Task Formulation}
Our approach centers on `episodic training', where each episode simulates a few-shot scenario, aligning training tasks closely with the conditions anticipated during testing. In this context, each `task' comprises inputs and expected outputs that the model must learn to recognize. The query set contains 10 random samples from the dataset for our experiments, evaluated under 1-shot, 5-shot, and 10-shot scenarios. The training involved 500 tasks per epoch, while validation employed 100 tasks. Instead of batch training, we opted for episodic training to prepare the model for real-world few-shot tasks more effectively.

To rigorously evaluate the model's robustness, we conducted both monolingual and cross-lingual experiments. Cross-lingual learning involves training on the script of one language and testing on another, while monolingual learning involves both training and testing on the script of the same language. Our experimental framework consisted of four evaluation methods:

\subsubsection{Monolingual Intra-Dataset Evaluation}
This method involved training and testing on the same dataset but with different character classes. We trained on consonants (39 classes) and tested on vowels (11 classes) from the CMATERdb 3.1.2 dataset\cite{0007md__mostafizur_rahman_2023_cmaterdb}. This approach evaluated the model’s ability to generalize within the same dataset to unseen character types.

\subsubsection{Monolingual Inter-Dataset Evaluation}    
This evaluation assessed the model’s capability to generalize to a different dataset with similar but distinct data distributions. In this method, we trained the model on the BanglaLekha-Isolated dataset's consonants and numerals (49 classes) \cite{0006banglalekhaisolated}  and tested it on the CMATERdb 3.1.2 dataset's vowels (11 classes). 

\subsubsection{Cross-Lingual Transfer}
To evaluate cross-script generalization, we trained on the Devanagari dataset (46 classes) \cite{0024Devanagari} and tested on the Bengali CMATERdb 3.1.2 (50 classes). This method tested the model's ability to transfer knowledge learned from one script (Devanagari) to another (Bengali).

\subsubsection{Split Digit Testing} 
This approach was designed to evaluate the model’s generalization ability within numerals. For this method, we used the NumtaDB dataset \cite{0015numtadb}, training on digits 0-5 (6 classes) and testing on digits 6-9 (4 classes). 

This multi-faceted evaluation strategy enables a thorough assessment of our FSL model’s versatility and robustness across varying conditions and data sources, underscoring its potential for practical applications in handwritten character recognition across diverse linguistic contexts.

\begin{figure*}[tbp]
    \centering
    \includegraphics[width=.915\textwidth]{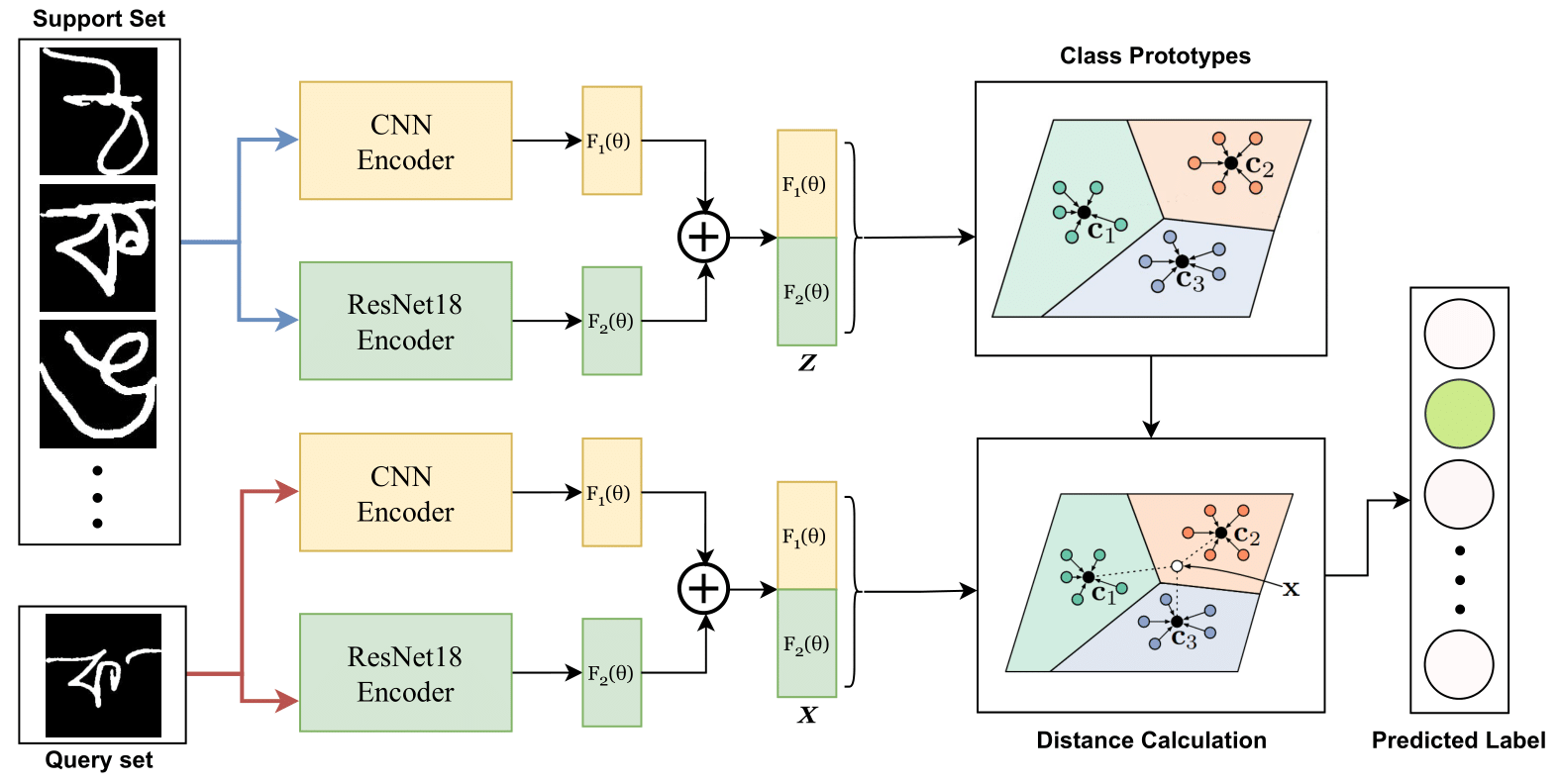}
    \caption{\textbf{Overview of our proposed SynergiProtonet}. The architecture consists of a dual-encoder system, integrating CNN and ResNet18 encoders, to process both Support Set and Query Set samples. Each image is passed through both encoders, yielding feature vectors \( F_1(\theta) \) from the CNN and \( F_2(\theta) \) from the ResNet18. These feature vectors are concatenated to form a combined representation \( Z \) for each Support image, and a Query feature vector \( X \) for the Query image. Class prototypes \( C_i \) are calculated by averaging the combined feature vectors of each class in the Support Set. For classification, the Euclidean distance is computed between \( X \) and each prototype \( C_i \), with a softmax function applied to the distances to yield class probabilities, allowing the model to classify the Query image based on its proximity to class prototypes in the feature space.}

    \label{fig:archi}    
\end{figure*}

% \begin{figure*}[tbp]
%     \centering
%     \includegraphics[width=0.88\textwidth]{images/synerginetModel.png}
%     \caption{Few-shot classifier architecture using our proposed network}
%     \label{fig:archi}    
% \end{figure*}
% \setlength{\abovecaptionskip}{0pt}

\subsection{Few Shot Learning Approaches}

\subsubsection{Matching Network} This network introduces an attention-based framework for FSL, enabling efficient comparison between query and support images \cite{0009Vinyals_Blundell_Lillicrap_Kavukcuoglu_Wierstra_2016}. Images are initially embedded into a feature space via a CNN to produce feature vectors, and an attention mechanism calculates similarity scores between the query and support images using a similarity metric, typically cosine similarity. Classification of the query image is performed by aggregating the support labels weighted by these similarity scores, allowing for effective label propagation across limited samples.

\subsubsection{Relation Network} The Relation Network framework enables flexible few-shot classification by learning a deep, trainable distance metric for comparing query images to support samples \cite{0011relationnetwork}. The images are embedded into a feature space through a CNN, and embeddings from the support and query sets are concatenated before being processed by a relation module that outputs a relation score, representing similarity. In $k$-shot settings, support embeddings are aggregated to form a class-level feature map for each class, which is then compared to the query. Training employs mean squared error (MSE) loss, with matched pairs labeled as 1 and mismatched pairs as 0, allowing the model to optimize for accurate similarity learning. During inference, the query is classified based on the highest relation score among the support samples.

\subsubsection{BD-CSPN (Bias Diminishing Cosine Similarity Prototypical Network)} 
BD-CSPN  advances FSL by refining class prototypes to mitigate intra- and inter-class biases, improving classification accuracy and robustness \cite{0010bdcspn}. This approach begins by extracting features with a CNN and generating basic prototypes as the average of support set embeddings per class. To address intra-class bias, BD-CSPN incorporates high-confidence query samples into the support set via pseudo-labeling, recalculating prototypes as a weighted average. For cross-class bias, it shifts the query embeddings towards support set prototypes, refining classification through cosine similarity with the nearest adjusted prototype.

\subsubsection{SimpleShot} The SimpleShot architecture is a lightweight approach to few-shot classification based on nearest-neighbor principles \cite{0008wang2019simpleshotrevisitingnearestneighborclassification}. After feature extraction with a CNN, SimpleShot applies centering by subtracting the mean feature vector of base classes and normalizes embeddings to unit norm using L2 normalization. For classification, it employs a nearest-neighbor classifier with Euclidean distance. In one-shot settings, the query is classified based on the closest support sample, while in multi-shot cases, class centroids are derived from averaged support features, enhancing classification efficiency.

\subsubsection{Prototypical Network} Prototypical Networks represent each class through a prototype computed as the mean embedding of the support set  \cite{0012Snell_Swersky_Zemel_2017b}. A CNN transforms images into high-dimensional vectors, with prototypes serving as centroids in embedding space. Classification is performed by calculating the Euclidean distance between the query embedding and each prototype and assigning the query to the nearest prototype. A softmax function transforms these distances into probabilities, while log-softmax loss penalizes misclassifications during training, allowing efficient learning from minimal data and rendering Prototypical Networks particularly well-suited to data-scarce environments.

\subsection{SynergiProtoNet}
Our proposed architecture, `SynergiProtoNet' integrates both a CNN-based encoder and a ResNet18-based encoder, as shown in Fig.~\ref{fig:archi}. It is designed to capture a rich spectrum of features for robust few-shot classification. 

The CNN Encoder is designed to capture fine-grained, low-level features critical for differentiating subtle patterns in handwritten characters. The first two layers employ $3\times3$ convolutional layers with 64 filters, followed by Batch Normalization (momentum=1), ReLU activation, and $2\times2$ Max Pooling. These layers are optimized to detect foundational image elements, including edges, textures, and simple shapes. Layers 3 \& 4 utilize additional $3\times3$ convolutional layers (64 filters, padding=1) with Batch Normalization and ReLU activation. These layers extend the receptive field, facilitating the extraction of intermediate-level features such as contours, corners, and detailed textural patterns.

The ResNet encoder utilizes a pre-trained ResNet18 model, focusing on extracting higher-level features by leveraging deep network architecture. Residual Connections mitigate the vanishing gradient problem, allowing the network to learn deeper, more abstract features without degradation. Deep Layers build increasingly complex feature hierarchies, extracting high-level features such as complex textures and object configurations essential for distinguishing between broader classes.

The combined encoder shown in Fig.~\ref{fig:archi} merges the outputs of the CNN and ResNet18 encoders to create a comprehensive feature representation that leverages both detailed local features and abstract global features. The outputs from both encoders are concatenated along the feature dimension, resulting in a combined feature vector that encapsulates a wide range of spatial information. This includes low-level features like edges, textures, corners, and fine patterns from the CNN Encoder; along with high-level features like complex textures and object configurations from the ResNet18 Encoder. This hybrid approach ensures that the feature representation is rich and diverse, capturing both intricate details and broader context, enhancing the model's ability to accurately classify few-shot learning tasks. The final output is flattened into a 1-dimensional vector, preserving detailed spatial information essential for distinguishing subtle differences between classes.

The few-shot classifier, implemented using Prototypical Networks, operates by computing class prototypes from the combined feature embeddings of the support set and classifying query samples based on their proximity to these prototypes. The embeddings are created by concatenating the outputs of the CNN and ResNet18 encoders along the feature dimension, resulting in a combined feature vector $z$ that encapsulates both low-level and high-level features (\equationautorefname~\ref{eq:concat}).
\begin{equation}
z = \text{concatenate}(f_{\text{CNN}}(x), f_{\text{ResNet}}(x), \text{dim}=1)
\label{eq:concat}
\end{equation}
where \( f_{\text{CNN}} \) and \( f_{\text{ResNet}} \) are the feature extraction functions of the CNN and ResNet18 encoders, respectively.

For each class $c$ in the support set $S$, the prototype $p_c$ is computed as the mean vector of the embedded support examples \(\{z_i\}_{i=1}^{K}\), where \( K \) is the number of support examples per class, as shown in (\equationautorefname~\ref{eq:prototype}).
\begin{equation}
p_c = \frac{1}{K} \sum_{i=1}^{K} f_\phi(x_i^{(c)})
\label{eq:prototype}
\end{equation}

To classify a query image $x_q$, the Euclidean distance between its embedding $ f_\phi(x_q)$ and each class prototype $p_c$ is computed (\equationautorefname~\ref{eq:distance}).
\begin{equation}
d(z_q, p_c) = \| z_q - p_c \|_2
\label{eq:distance}
\end{equation}

Finally, a softmax function is applied over the negative distances to convert them into probabilities (\equationautorefname~\ref{eq:probability}).
\begin{equation}
p_\phi(y = c \mid z_q) = \frac{\exp(-d(z_q, p_c))}{\sum_{c'=1}^{N_c} \exp(-d(z_q, p_{c'}))}
\label{eq:probability}
\end{equation}

During training, the log-softmax loss is used to penalize the model when it fails to predict the correct class, driving the backpropagation process to minimize classification errors. This overall training process enables SynergiProtoNet to generalize effectively from limited examples, achieving high accuracy in challenging few-shot classification tasks.

\begin{table*}[t]
% \centering
% \caption{Monolingual Performance Analysis}

\begin{minipage}{0.5\linewidth}
\centering
\caption{Monolingual Intra-Dataset Evaluation}
\label{tab:mono-intra-table}
\resizebox{\textwidth}{!}{%
\begin{tabular}{L{2.7cm} C{1.2cm} C{0.5cm} | C{1.2cm} C{0.5cm} | C{1.2cm} C{0.5cm}}
\toprule
\multirow{2}{*}{\textbf{Network}} & \multicolumn{2}{C{1.1cm}}{\textbf{1-shot}} & \multicolumn{2}{C{1.1cm}}{\textbf{5-shot}} & \multicolumn{2}{C{1.1cm}}{\textbf{10-shot}} \vspace{1pt} \\ 
\cmidrule{2-7}
 & \textbf{Acc(\%)} & \textbf{F1} & \textbf{Acc(\%)} & \textbf{F1} & \textbf{Acc(\%)} & \textbf{F1} \\ 
\midrule

Matching \cite{0009Vinyals_Blundell_Lillicrap_Kavukcuoglu_Wierstra_2016} & 69.64 & 0.69 & 38.66 & 0.39 & 36.36 & 0.36 \\ 

Simpleshot \cite{0008wang2019simpleshotrevisitingnearestneighborclassification} & 69.42 & 0.7 & 75.68 & 0.78 & 82.9 & 0.82 \\ 

Relation \cite{0011relationnetwork} & 77.10 & 0.76 & 85.58 & 0.87 & 83.2 & 0.83 \\ 

BD-CSPN \cite{0010bdcspn} & 69.68 & 0.71 & 76.58 & 0.76 & 83.48 & 0.82 \\ 

Prototypical \cite{0012Snell_Swersky_Zemel_2017b} & 74.48 & 0.74 & 87.88 & 0.87 & 87.56 & 0.87 \\ 

% SynergiProtoNet \\(without Pretraining) & 44.5 & 0.45 & 65.16 & 0.66 & 68.76 & 0.68 \\ 

SynergiProtoNet (ours) & \textbf{79.1} & \textbf{0.79} & \textbf{88.95} & \textbf{0.88} & \textbf{90.04} & \textbf{0.9} \\

\bottomrule
\end{tabular}
}
\end{minipage}
\hfill
\begin{minipage}{0.5\linewidth}
\centering
\caption{Monolingual Inter-Dataset Evaluation}
\label{tab:mono-inter-table}
\resizebox{0.97\textwidth}{!}{%
\begin{tabular}{L{2.7cm} C{1.2cm} C{0.5cm} | C{1.2cm} C{0.5cm} | C{1.2cm} C{0.5cm}}
\toprule
    \multirow{2}{*}{\textbf{Network}}    & 
    \multicolumn{2}{C{1.1cm}}{\textbf{1-shot}} & 
    \multicolumn{2}{C{1.1cm}}{\textbf{5-shot}} & 
    \multicolumn{2}{C{1.1cm}}{\textbf{10-shot}} \vspace{1pt}\\ 
    \cmidrule{2-7} 
    & 
    \textbf{Acc(\%)} & \textbf{F1} & 
    \textbf{Acc(\%)} & \textbf{F1} & 
    \textbf{Acc(\%)} & \textbf{F1} \\ 
    \midrule

    Matching \cite{0009Vinyals_Blundell_Lillicrap_Kavukcuoglu_Wierstra_2016} & 
    48.6 & 0.52 & 
    41.8 & 0.39 & 
    26.9 & 0.26 \\ 

    Simpleshot \cite{0008wang2019simpleshotrevisitingnearestneighborclassification} & 
    51.34 & 0.55 & 
    63.54 & 0.63 & 
    64.94 & 0.66 \\

    Relation \cite{0011relationnetwork} & 
    56.46 & 0.58 & 
    74.02 & 0.72 & 
    65.04 & 0.67 \\ 
    
    BD-CSPN \cite{0010bdcspn} & 
    50.84 & 0.5 & 
    60.3 & 0.6 & 
    69.2 & 0.7 \\ 
    
    Prototypical \cite{0012Snell_Swersky_Zemel_2017b} & 
    54.28 & 0.55 & 
    76.18 & 0.75 & 
    76.24 & 0.75 \\ 
    
    % SynergiProtoNet \\ (without Pretraining) & 
    % 44.5 & 0.45 & 
    % 65.16 & 0.66 & 
    % 68.76 & 0.68 \\ 
    
    SynergiProtoNet (ours) & 
    \textbf{59.02} & \textbf{0.59} & 
    \textbf{77.68} & \textbf{0.78} & 
    \textbf{81.36} & \textbf{0.83} \\ 
 
    \bottomrule
\end{tabular}
}
\end{minipage}
\end{table*}

%For Bengali character and digit recognition, the CMATERdb 3.1.2\cite{0007md__mostafizur_rahman_2023_cmaterdb} and BengaLekha-Isolated\cite{0006banglalekhaisolated} datasets were divided with consonants and digits in the Train split and vowels in the Test split, excluding the Compound characters. 
%The NumtaDB dataset \cite{0015numtadb} consists of five subsets: a, b, c, d, and e. These subsets were combined and subsequently divided into training and testing sets. Digits 0-5 were placed in the Train folder, while digits 6-9 were allocated to the Test split. 
%The Devanagari\cite{0024Devanagari} dataset retained its original structure for cross-lingual experiments, with 1700 training and 300 test examples per character. Support sets were created by selecting a subset of images from each class, and 10 query images per class were preprocessed similarly. %These steps ensured consistent input data format, enabling effective training and evaluation of the FSL models across different datasets and scenarios.

\subsection{Training and Evaluation Procedures}
The model parameters were optimized using the SGD algorithm with a learning rate of $10^{-2}$, momentum of 0.9, and weight decay of $5 \times 10^{-4}$. To ensure uniformity and compatibility, images were resized to $84 \times 84 \times 3$ to satisfy the input requirements of the ResNet18 backbone.  
A Multi-Step scheduler adjusted the learning rate at specified milestones, reducing it by a factor of 0.1 for fine-tuning. Cross-entropy loss was used to measure the discrepancy between predicted and true labels, making it suitable for multi-class classification tasks. The training process spanned 30 epochs, with episodic training involving 500 tasks for training and 100 tasks for validation in each epoch. For Bangla characters, FSL was performed using a 5-way classification setup, while for digits, a 3-way classification setup was used. During each episode, the support set images and labels were used to update the model’s prototypes, which were then used to predict the classes of the query set images. Overlapping classes were avoided in all experiments to ensure fair evaluation.

% We implemented four distinct evaluation methods to rigorously test our few-shot learning model across various scenarios and datasets, demonstrating its robustness and versatility:

% \subsubsection{Monolingual Intra-Dataset Evaluation}
% This method involved training and testing on the same dataset but with different character types. Specifically, we trained on consonants and tested on vowels within the CMATERdb 3.1.2 dataset. This approach evaluated the model’s ability to generalize within the same dataset to unseen character types.

% \subsubsection{Monolingual Inter-Dataset Evaluation}
% In this method, we trained the model on the BanglaLekha-Isolated dataset and tested it on the CMATERdb 3.1.2 dataset. This evaluation assessed the model’s capability to generalize to a different dataset with similar but distinct data distributions.

\begin{table*}[t]
\centering
\begin{minipage}{.5\linewidth}
  \centering
  \caption{Cross-Lingual Performance Analysis}
  \label{tab:cross-table}
  \resizebox{0.97\textwidth}{!}{%
  \begin{tabular}{L{2.7cm} C{1.2cm} C{0.5cm} | C{1.2cm} C{0.5cm} | C{1.2cm} C{0.5cm}}
  \toprule
      \multirow{2}{*}{\textbf{Network}}    & 
      \multicolumn{2}{C{1.1cm}}{\textbf{1-shot}} & 
      \multicolumn{2}{C{1.1cm}}{\textbf{5-shot}} & 
      \multicolumn{2}{C{1.1cm}}{\textbf{10-shot}} \vspace{1pt}\\ 
      \cmidrule{2-7} 
      & 
      \textbf{Acc(\%)} & \textbf{F1} & 
      \textbf{Acc(\%)} & \textbf{F1} & 
      \textbf{Acc(\%)} & \textbf{F1} \\ 
      \midrule
  
      Matching \cite{0009Vinyals_Blundell_Lillicrap_Kavukcuoglu_Wierstra_2016} & 
      52.84 & 0.52 & 
      26.68 & 0.27 & 
      37.58 & 0.38 \\ 
  
      Simpleshot \cite{0008wang2019simpleshotrevisitingnearestneighborclassification} & 
      42.92 & 0.46 & 
      56.14 & 0.53 & 
      55.10 & 0.54 \\ 
      
      BD-CSPN \cite{0010bdcspn} & 
      44.58 & 0.45 & 
      54.72 & 0.52 & 
      57.72 & 0.57 \\ 
  
      Prototypical \cite{0012Snell_Swersky_Zemel_2017b} & 
      53.64 & 0.55 & 
      76.74 & 0.77 & 
      79.48 & 0.79 \\ 
      
      Relation \cite{0011relationnetwork} & 
      \textbf{61.12} & \textbf{0.61} & 
      74.02 & 0.72 & 
      81.93 & 0.82 \\ 

      % SynergiProtoNet \\ (without Pretraining) & 
      % 50.46 & 0.51 & 
      % 71.14 & 0.7 & 
      % 76.98 & 0.76 \\ 
  
      SynergiProtoNet (ours) & 
      58.59 & 0.55 & 
      \textbf{76.84} & \textbf{0.77} & 
      \textbf{82.12} & \textbf{0.82} \\ 
  
      \bottomrule
  \end{tabular}}
\end{minipage}%
\begin{minipage}{.5\linewidth}
  \centering
  \caption{Performance Analysis of Split Digit Testing}
  \label{tab:split-digit}
  \resizebox{0.97\textwidth}{!}{%
  \begin{tabular}{L{2.7cm} C{1.2cm} C{0.5cm} | C{1.2cm} C{0.5cm} | C{1.2cm} C{0.5cm}}
  \toprule
      \multirow{2}{*}{\textbf{Network}}    & 
      \multicolumn{2}{C{1.1cm}}{\textbf{1-shot}} & 
      \multicolumn{2}{C{1.1cm}}{\textbf{5-shot}} & 
      \multicolumn{2}{C{1.1cm}}{\textbf{10-shot}} \vspace{1pt}\\ 
      \cmidrule{2-7}
      & 
      \textbf{Acc(\%)} & \textbf{F1} & 
      \textbf{Acc(\%)} & \textbf{F1} & 
      \textbf{Acc(\%)} & \textbf{F1} \\ 
      \midrule
  
      Matching \cite{0009Vinyals_Blundell_Lillicrap_Kavukcuoglu_Wierstra_2016} & 
      \textbf{73.67} & \textbf{0.74} & 
      33.1 & 0.29 & 
      32.17 & 0.33 \\

      BD-CSPN \cite{0010bdcspn} & 
      55.43 & 0.59 & 
      64.43 & 0.64 & 
      69.17 & 0.68 \\

      Simpleshot \cite{0008wang2019simpleshotrevisitingnearestneighborclassification} & 
      65.6 & 0.61 & 
      72.9 & 0.7 & 
      76.97 & 0.76 \\

      Relation \cite{0011relationnetwork} & 
      64.97 & 0.63 & 
      79.83 & 0.8 & 
      81.37 & 0.83 \\

      Prototypical \cite{0012Snell_Swersky_Zemel_2017b} & 
      66.57 & 0.65 & 
      76.33 & 0.75 & 
      87.1 & 0.86 \\

      % SynergiProtoNet \\ (without Pretraining) & 
      % 46.47 & 0.48 & 
      % 52.6 & 0.53 & 
      % 57.6 & 0.57 \\ 

      SynergiProtoNet (ours) & 
      37.4 & 0.37 & 
      \textbf{83.73} & \textbf{0.83} & 
      \textbf{88.3} & \textbf{0.87} \\

      \bottomrule
  \end{tabular}}
\end{minipage}
\end{table*}

% \subsubsection{Cross-Lingual Transfer}
% Here, we trained the model on the Devanagari dataset and tested it on the CMATERdb 3.1.2 dataset in 1, 5, 10-shot scenarios. This method tested the model's ability to transfer knowledge learned from one script (Devanagari) to another (Bengali), highlighting its cross-lingual generalization capabilities.

% \subsubsection{Split Digit Testing}
% For this method, we used the NumtaDB dataset, training on digits 0-5 and testing on digits 6-9. This approach was designed to evaluate the model’s performance on completely unseen classes within the same domain, ensuring it could handle variations within the same type of data.

% This comprehensive approach ensured rigorous testing of our few-shot learning model, demonstrating its robustness and versatility across various scenarios and datasets.

\section{Result Analysis} 
SynergiProtoNet, with its hybrid architecture, leverages a CNN for extracting lower-level local features and a pre-trained ResNet18 for capturing high-level global features. This fusion of detailed and abstract feature extraction facilitates comprehensive representation, enabling the model to perform consistently across diverse tasks and datasets.

% Monolingual Intra-Dataset Performance (\tableautorefname~\ref{tab:mono-intra-table}) demonstrates SynergiProtoNet's superior performance on the CMATERdb 3.1.2 dataset, achieving the highest accuracy and F1 score across all shot scenarios. This strong performance in intra-dataset evaluation highlights the model's ability to recognize complex handwriting variations within the same dataset, effectively capturing nuanced differences between the different characters. In contrast, other models couldn't capture complex handwriting variations as much as SynergiProtoNet because our model is deeper.

\subsubsection{Monolingual Intra-Dataset Performance}
\tableautorefname~\ref{tab:mono-intra-table}  reveals SynergiProtoNet's capacity on the CMATERdb 3.1.2 dataset, where it achieves the highest accuracy and F1 score across all shot scenarios. This impressive performance underscores the model's adeptness in handling the intricate and varied forms of handwriting that can exist within a single dataset, successfully distinguishing complex patterns and subtle distinctions between individual characters. SynergiProtoNet's architecture, meticulously designed to balance depth with precision, enables it to capture even the most nuanced variations in character formation, which are often missed by other models. While competing models struggle to maintain accuracy when faced with complex handwriting differences, SynergiProtoNet’s deeper and more sophisticated structure empowers it to process and learn from these detailed features, consistently delivering high-precision recognition. This capability is particularly vital in real-world applications where script recognition needs to accommodate a high degree of variability without compromising on accuracy. By setting itself apart with such refined intra-dataset recognition, SynergiProtoNet not only demonstrates its robustness in controlled scenarios but also suggests promising adaptability to real-world complexities in handwritten character recognition across different settings and styles.

\subsubsection{Monolingual Inter-Dataset Performance}
\tableautorefname~\ref{tab:mono-inter-table} demonstrates SynergiProtoNet’s robustness when trained on one dataset containing Bangla handwritten scripts and evaluated on a different one, where the training and testing sets contained entirely distinct classes of the same language. The superior performance of SynergiProtoNet in this setting can be attributed to its enhanced encoder, which initially captures intricate, task-relevant features, subsequently refined by ResNet18's deep feature extraction. This layered feature capture enables better generalization across datasets, adapting to the inherent diversity of Bangla handwritten characters more effectively than other models. In contrast, models such as Prototypical Networks and Relation Networks, while performing reasonably well, lack sophisticated feature extraction capabilities, leading to slightly lower performance metrics. Similarly, the performance of Simpleshot, BD-CSPN and Matching Networks, still lags behind SynergiProtoNet, underscoring due to similar reasons. The rigorous testing through this evaluation highlights SynergiProtoNet's versatility.

\subsubsection{Cross-Lingual Performance Analysis}
\tableautorefname~\ref{tab:cross-table} explores the model's performance in cross-lingual scenarios, where it was trained on Devanagari characters and tested on Bengali characters. Here, SynergiProtoNet demonstrated strong performance in the 5-shot and 10-shot scenarios, maintaining balanced accuracy across multiple shots. Although the Relation Network achieved marginally higher accuracy in the 1-shot scenario, it lagged behind SynergiProtoNet in multi-shot settings. This consistent performance across shots illustrates SynergiProtoNet's ability to manage the cross-lingual shift effectively, even as other models struggle to generalize across the language barrier. Prototypical Networks perform reasonably well but have slightly lower generalization capacity compared to ours. SimpleShot architecture, relying on nearest-neighbor classification, performed lowest overall due to its limited feature extraction depth, unable to capture complex, language-specific nuances. BD-CSPN and Matching Networks show lower performance, particularly in the 5-shot scenario for Matching Networks, indicating a potential overfitting issue where the models fail to generalize from the provided examples. While SynergiProtoNet's performance in 1-shot is slightly diminished due to its architectural complexity which limits its trainability with such limited data, its results were on par with the other models. Its steady performance across other shots showcases its robustness in cross-lingual scenarios.

\subsubsection{Split Digit Testing}
\tableautorefname~\ref{tab:split-digit} evaluated the model on NumtaDB, a dataset featuring diverse Bengali numerals across five different subsets, each varying in style and format (e.g., handwritten in pencil, pen, or generated digitally on contrasting backgrounds). Although the Matching Network performed well in the 1-shot setting, SynergiProtoNet demonstrated significantly better accuracy in the 5-shot and 10-shot scenarios. The diversity in NumtaDB’s digit styles presented a challenge for many models, with SynergiProtoNet’s hybrid architecture showing resilience and adaptability in capturing the underlying structures across varied numeral styles. The model's performance illustrates its robustness, effectively handling both simple and complex numeral presentations, while other methods struggled to generalize from the diverse data.

In summary, these evaluations reveal that SynergiProtoNet’s hybrid architecture is well-suited for handling intricate details and abstract representations, achieving top performance across monolingual, cross-lingual, and complex numeral datasets. This versatility positions SynergiProtoNet as a powerful solution for few-shot learning tasks in complex, diverse handwriting recognition applications.

\section{Conclusion} 

This study provides a thorough performance analysis of several state-of-the-art few-shot learning models in Bangla handwritten character and digit recognition under diversified circumstances. Moreover, we introduce SynergiProtoNet which is specifically designed to tackle the complexities of Bangla handwritten scripts with limited data. By employing a hybrid encoder structure that combines CNN-based local feature extraction with high-level representations from ResNet18, SynergiProtoNet captures a rich spectrum of visual details and contextual patterns, demonstrating its effectiveness across varied monolingual and cross-lingual scenarios. Experimental results indicate that SynergiProtoNet not only achieves state-of-the-art accuracy within Bangla datasets but also adapts effectively to cross-dataset evaluations, highlighting its resilience to inter-dataset variability and class diversity in handwritten scripts. Future research could improve SynergiProtoNet to support compound and non-Latin characters, enhancing its versatility across various scripts. Further refinement of its hybrid architecture with better attention mechanisms or adaptive feature fusion may increase robustness in diverse handwriting scenarios. This positions SynergiProtoNet as an effective solution for industries, promoting sustainable technology by minimizing continuous data acquisition and retraining.

\bibliographystyle{IEEEtran}
\bibliography{citations}

\end{document}